\documentclass[sigconf]{acmart}
\AtBeginDocument{
  }

\copyrightyear{2026}
\acmYear{2026}
\setcopyright{cc}
\setcctype{by-nc-nd}
\acmConference[ICMR '26]{International Conference on Multimedia Retrieval}{June 16--19, 2026}{Amsterdam, Netherlands}
\acmBooktitle{International Conference on Multimedia Retrieval (ICMR '26), June 16--19, 2026, Amsterdam, Netherlands}
\acmDOI{10.1145/3805622.3810759}
\acmISBN{979-8-4007-2617-0/2026/06}

\usepackage{multirow}
\usepackage{algorithm}
\usepackage{algorithmic}
\usepackage{subfigure}
\usepackage{enumitem}

\usepackage{amssymb}
\usepackage{makecell}

\begin{document}

\title{Towards Generalized Image Manipulation Localization via Score-based Model}

\author{Yunfei Wang}
\authornote{Both authors contributed equally to this research.}
\affiliation{
  \institution{Sichuan University}
  \city{Chengdu}
  \country{China}
}
\email{2024223045097@stu.scu.edu.cn}

\author{Bo Du}
\authornotemark[1]
\authornote{Corresponding authors.}
\affiliation{
  \institution{Sichuan University}
  \city{Chengdu}
  \country{China}
}
\email{2024223045105@stu.scu.edu.cn}

\author{Zhe Yang}
\affiliation{
  \institution{Sichuan University}
  \city{Chengdu}
  \country{China}
}
\email{yangzhe@stu.scu.edu.cn}

\author{Xin Liu}
\affiliation{
  \institution{Sichuan University}
  \city{Chengdu}
  \country{China}
}
\email{lovasleeping@gmail.com}

\author{Zhiyu Lin}
\affiliation{
  \institution{Sichuan University}
  \city{Chengdu}
  \country{China}
}
\email{zhiyulin@stu.scu.edu.cn}

\author{Tianxin Xu}
\affiliation{
  \institution{Sichuan University}
  \city{Chengdu}
  \country{China}
}
\email{xtx@stu.scu.edu.cn}

\author{Ji-Zhe Zhou}
\authornotemark[2]
\affiliation{
  \institution{Sichuan University}
  \city{Chengdu}
  \country{China}
}
\email{jzzhou@scu.edu.cn}

\begin{abstract}
With the rapid evolution of synthetic media, Image Manipulation Localization (IML) has emerged as a critical component in multimedia forensics for ensuring the integrity of digital content. However, generalization remains a core challenge, as existing discriminative methods typically learn a fixed decision boundary that tends to overfit to specific training artifacts and fails to adapt to unseen manipulation types. To address this, we propose \textit{DiffIML}, a novel framework that introduces score-based generative modeling to IML. Diverging from the direct estimation of hard boundaries, DiffIML approximates the score function, the gradient of the log-likelihood, to capture the intrinsic geometric topology of mask distributions. This paradigm leverages structural priors to iteratively recover coherent masks from noise, thereby circumventing the brittleness associated with discriminative models. 
Under this formulation, diffusion models serve as an effective numerical solver for the learned score function.
To ensure practicality, we respectively resolve the \textit{efficiency} and \textit{stability} bottlenecks of standard diffusion by: (1) utilizing a Lightweight Mask-Specific VAE for fast latent-space process and a decoupled architecture with a lightweight denoising UNet, (2) \textit{edge supervision} and \textit{error prior} to mitigate error accumulation during sampling. Extensive experiments of two distinct protocols on eight non-generative and three generative benchmarks demonstrate that DiffIML consistently outperforms state-of-the-art methods, yielding remarkable generalization improvements on diverse unseen datasets. The code is publicly available at \url{https://github.com/scu-zjz/DiffIML}.
\end{abstract} 



\maketitle

\section{Introduction}
\label{sec:intro}

With the rapid evolution of synthetic media, Image Manipulation Localization (IML) has become a critical component in multimedia forensics to ensure the integrity of digital media content. Specifically, IML aims to classify each pixel in an image as either manipulated or authentic, where "manipulation" is defined as the partial modification of a real image that yields semantic
discrepancies~\cite{ma2025imdl}. However, generalization remains a core challenge, as image manipulation techniques are constantly evolving (e.g., recent editing manipulation by generative models~\cite{stablediff, ramesh2021zero}), while public datasets cover only a limited range of manipulation types.

\begin{figure}[t]
  \centering
\includegraphics[width=0.95\columnwidth]{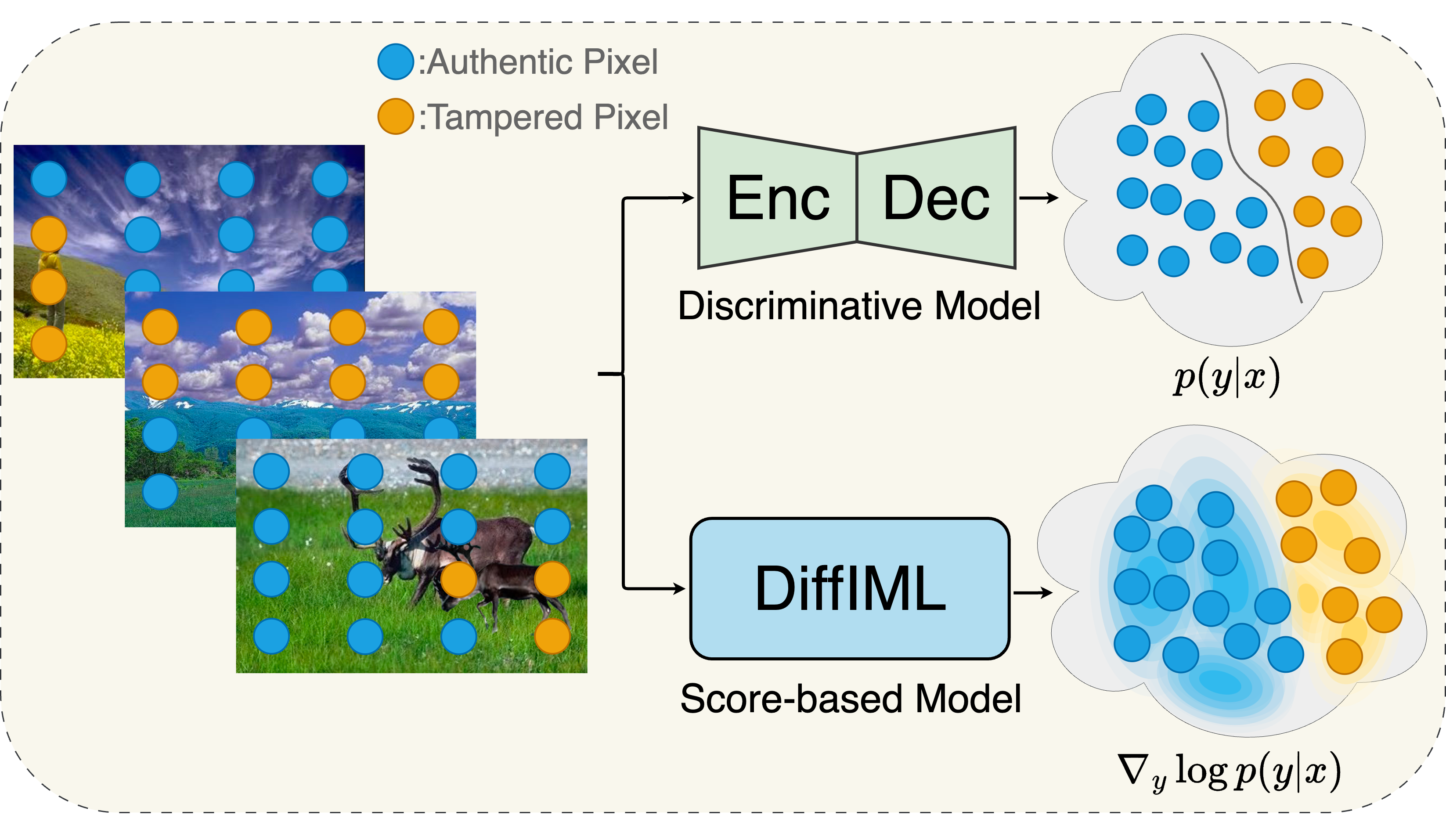}
  \caption{Conceptual comparison between Discriminative and Score-based IML.
\textit{Top}: Previous discriminative models learn a fixed decision boundary $p(y|x)$, which tends to overfit to the training distribution and fails on unseen manipulation types (sparse regions).
\textit{Bottom}: Our proposed DiffIML learns the score function $\nabla_y \log p(y|x)$ (visualized as vector fields). By learning the gradient direction towards the data distribution, DiffIML captures the generalized geometric structure of valid masks, allowing it to accurately infer masks for unseen manipulations via iterative refinement (Langevin dynamics).}
\Description{A conceptual comparison of two paradigms for image manipulation localization. The top part illustrates discriminative methods as learning a fixed decision boundary that fits seen manipulation types but fails on sparse unseen regions. The bottom part illustrates the proposed score-based method as learning a vector field of score gradients that guides samples toward the mask distribution, enabling iterative refinement and better generalization to unseen manipulations.}
  \label{fig:dis_gen}
\end{figure}

For a training set $\{(x_i, y_i)\}^N_{i=1}$, where $x$ is the image and $y$ is the pixel-level binary mask, previous methods predominantly model the conditional distribution $p(y|x)$ directly using discriminative encoder-decoder architectures. As illustrated in Figure \ref{fig:dis_gen} (top), these methods aim to learn a definitive decision boundary to separate manipulated pixels from authentic ones. Although effective on seen data, such deterministic mapping suffers from intrinsic generalization limitations. Since training data cannot exhaust all manipulation types, the learned boundary often overfits to the specific artifacts (tampering traces) of the training set. When facing unseen manipulations (sparse regions in the distribution), the fixed decision boundary fails to adapt, leading to misclassification~\cite{ma2025imdl}.

\begin{figure*}[t]
  \centering
\includegraphics[width=0.88\linewidth]{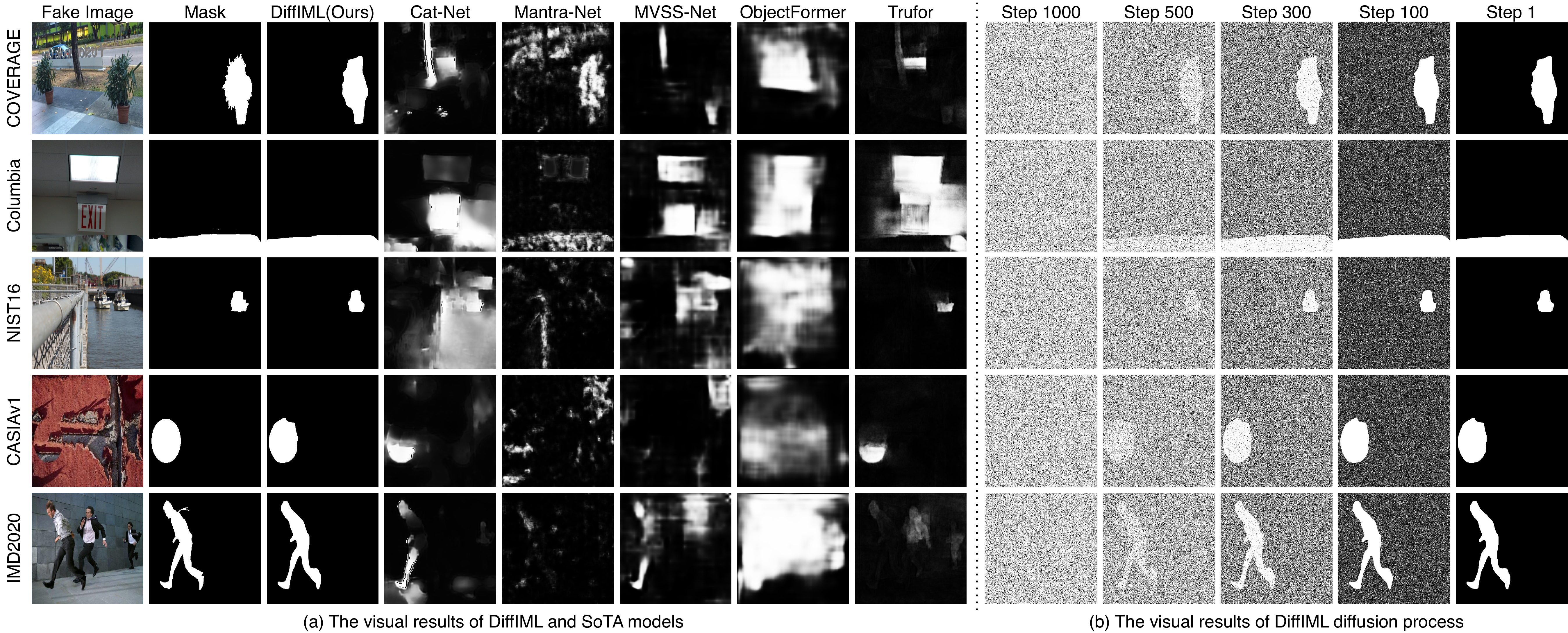}
  \caption{(a) Comparison of DiffIML and discriminative baselines: DiffIML preserves spatial coherence and structural completeness of masks on unseen fake images, while discriminative baselines exhibit severe mask fragmentation. (b) Samples in the diffusion process: Gradually transitioning from Gaussian noise to pure mask from left to right.} 
  \Description{A two-part figure. Part (a) compares visual mask predictions of DiffIML and several discriminative baselines on unseen manipulated images from multiple datasets. Each row shows a fake image, the ground-truth mask, the prediction of DiffIML, and the predictions of baseline methods. DiffIML produces masks that are more complete, coherent, and closer to the ground truth, while the baselines often produce fragmented, blurry, or incomplete masks. Part (b) shows the diffusion denoising process over multiple steps, where the predicted mask gradually evolves from noisy initialization to a clear binary mask from left to right.}
  \label{fig:com_sota_pred}
\end{figure*}

To address this, we introduce the score-based generative modeling framework to IML, termed \textbf{DiffIML}. Diverging from the direct estimation of hard decision boundaries, score-based models approximate the score function, defined as the gradient of the log-likelihood $\nabla_y \log p(y|x)$. As illustrated in Figure \ref{fig:dis_gen} (bottom), this score manifests as a vector field that guides the sampling trajectory toward high-density regions of the ground-truth distribution. Fundamentally, this paradigm fosters superior generalization by shifting the learning objective: instead of overfitting to specific, variable manipulation artifacts, the score function captures the intrinsic geometric topology of the mask distribution. This mechanism effectively models the iterative recovery of coherent masks from noise, conditioned on the image context.

Consequently, by leveraging learned structural priors, such as spatial smoothness and semantic boundary alignment, the score-based model can accurately infer masks for unseen manipulation types, thereby circumventing the brittleness associated with rigid classification boundaries of discriminative models. This is empirically verified in Fig. \ref{fig:com_sota_pred}(a), where discriminative baselines exhibit severe mask fragmentation on unseen images due to their inflexible boundaries. In contrast, DiffIML preserves spatial coherence and structural completeness, demonstrating that the score-based prior effectively guides the recovery of accurate masks even in the presence of novel manipulation traces. Notably, our experiments (Sec. \ref{sec:diff_paradigm}) demonstrate that simply switching from a discriminative paradigm to a score-based diffusion paradigm on the same backbone yields a remarkable generalization improvement, gaining an F1 score increase from 0.457 to 0.533.

Under the score-based generative formulation, we implement DiffIML using Denoising Diffusion Probabilistic Models (DDPMs)~\cite{ddpm, song2020score}, which effectively estimate the score function via a sequence of denoising autoencoders. However, a naive or direct instantiation of standard diffusion models for IML presents two critical bottlenecks: \textit{efficiency} and \textit{stability}. First, the \textit{efficiency bottleneck} arises from operating diffusion in the high-dimensional pixel space, which requires hundreds of iterative denoising steps. This leads to prohibitive computational costs and inference latency, making it impractical for real-world deployment. Second, the \textit{stability bottleneck} stems from the iterative sampling process (i.e., Langevin dynamics, Fig.~\ref{fig:com_sota_pred}(b)), which is prone to error accumulation, where small deviations in each step propagate and degrade the accuracy of the final predicted mask.

To tackle the first efficiency bottleneck, we propose a dual-efficiency strategy that improves efficiency from both the representation and model perspectives. First, we utilize a Lightweight Mask-Specific VAE~\cite{vae} by distilling the VAE of Stable Diffusion~\cite{stablediff} to compress the mask into a compact latent representation, which enables a fast, low-dimensional diffusion process. Second, we decouple the condition learning from the iterative denoising process. Specifically, we employ a dedicated learnable \textit{Condition Encoder} to thoroughly capture artifacts, which allows the computationally expensive iterative denoising to be executed by a significantly lightweight UNet~\cite{unet}. For the second stability bottleneck, we analyze two core causes of error accumulation in the iterative sampling process: \textit{prediction error} and \textit{iterative error} (detailed in Sec. \ref{proble_form}). We further propose \textit{edge supervision} and \textit{error prior} to mitigate them separately.

Our contributions are threefold:
\begin{itemize}
\item \textbf{New Perspective of Score-based Modeling for IML Generalization.} We argue that by learning the gradient of the log-likelihood rather than a fixed decision boundary, score-based models capture the intrinsic geometric structure of manipulation masks, enabling superior generalization to unseen manipulation types compared to discriminative baselines.
\item \textbf{Novel Efficient and Stable Architecture.} We propose \textit{DiffIML}, a practical framework designed for IML that resolves the efficiency and stability bottlenecks of standard score-based diffusion through latent-space modeling and error mitigation mechanisms.
\item \textbf{Superior Generalization through Extensive Experiments.} We conduct comprehensive experiments of two distinct protocols on eight non-generative and three generative benchmarks and demonstrate that \textit{DiffIML} consistently outperforms 8 state-of-the-art methods.

\end{itemize}
\section{Related Work}
\label{sec:related_work}
\begin{figure*}[t]
  \centering
  \includegraphics[width=0.8\linewidth]{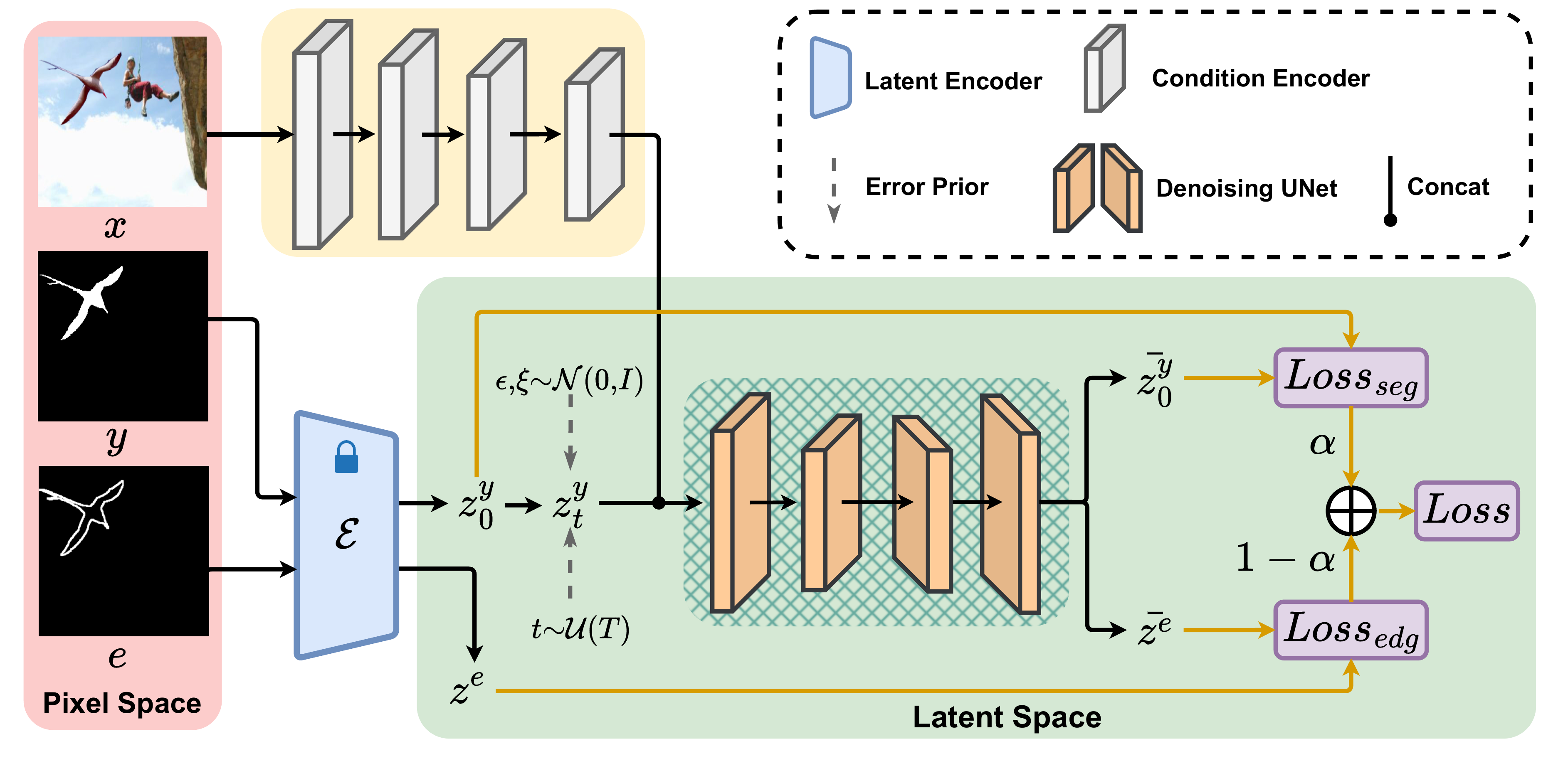}
  \caption{\textbf{Training architecture of the proposed DiffIML.} The condition encoder extracts the trainable artifact condition within the image. The mask and edge map transform into the latent representation through the latent encoder. The Langevin dynamics performs on the latent mask, and the latent edge map provides edge supervision.} 
  \Description{The training architecture of DiffIML. An input image is processed by a condition encoder to extract artifact-related condition features. The ground-truth mask and its corresponding edge map are separately encoded by a latent encoder into latent representations. These latent features are concatenated and used in the diffusion training process. A denoising U-Net predicts the latent outputs under the guidance of the image condition, while an error prior is introduced to improve stability during iterative denoising. The latent edge representation provides auxiliary edge supervision, and the model is trained to reconstruct both mask and edge information in latent space.}
  \label{fig:network}
\end{figure*}
\textbf{Image Manipulation Localization.} When an image is manipulated, it inevitably leaves manipulation traces (artifacts). Therefore, the key to the IML task is to locate the manipulated regions by detecting artifacts. Previous methods propose various extractors to capture artifacts beyond RGB domain. \citet{zhou2018learning} firstly propose an extra noise stream in the area of image manipulation localization, which leverages the noise features extracted from a Steganalysis Rich Model (SRM) layer~\cite{fridrich2012rich} to discover the noise inconsistency between authentic and tampered regions. \citet{wu2019mantra}, and \citet{hu2020span} use both SRM filters and BayerConv for noise artifacts extraction. \citet{wang2022objectformer} extracts frequency artifacts by transforming from RGB domain using Discrete Cosine Transform(DCT) and a high pass filter to provide complementary clues. Although these methods capture some artifacts beyond the RGB domain, likelihood-based learning approaches inherently tend to overfit the artifacts of manipulation types present in the training set, leading to poor generalization.

\textbf{Score-based Model.} Score-based models~\cite{song2019generative,ddpm} are a class of models that estimate the gradient of a distribution by score matching, originally proposed to avoid evaluating the normalizing constant. These models produce samples using Langevin dynamics, which approximately works by gradually moving a random initial sample to high density regions using the estimated gradient~\cite{song2020score}. The most effective recent score-based models are diffusion models~\cite{ddim, ddpm}, which have achieved outstanding performance in image and video generation. These models have shown great potential in unexplored applications such as IML, as they are able to capture the more generalized geometric properties of the distribution.

\section{Method}

\subsection{Problem Formulation}
\label{proble_form}

The objective of IML is to predict the manipulated pixels, alternatively referred to as the mask $y \in \mathbb{R}^{W \times H}$, in the input image $x \in \mathbb{R}^{W \times H \times 3}$. We adopt a score-based generative formulation for IML and instantiate it with a diffusion model to estimate the mask distribution. In addition to the binary mask $y$, we further derive a tampering edge map $e$ and model the joint distribution in a shared latent space for better boundary localization.

The training process (forward process) and sampling process (reverse process) of DiffIML are distinct. In the forward process, DiffIML defines a sequence of positive noise scales $\{\beta_1, \cdots, \beta_T\}$ with $T$ steps. For each data pair $(x,y)$, a discrete Markov chain $\{y_0,y_1,\cdots,y_T\}$ is constructed. The transition and the resulting marginal distribution are defined as:
\begin{align}
    p(y_t|y_{t-1}) &= \mathcal{N}(y_t; \sqrt{1-\beta_t}y_{t-1}, \beta_t I), \\
    q(y_t|y_0) &= \mathcal{N}(y_t; \sqrt{\bar{\alpha}_t}y_0, (1-\bar{\alpha}_t)I),
\end{align}
where $\bar{\alpha}_t:=\prod_{s=1}^t (1-\beta_s)$ and $y_0 = y$. The noise scales are pre-designed to ensure that $y_T$ approaches an isotropic Gaussian distribution $\mathcal{N}(0, I)$.

In the reverse process, a variational Markov chain is parameterized with:
\begin{equation}
    p_{\theta}(y_{t-1}|y_t)=\mathcal{N}(y_{t-1};\frac{1}{\sqrt{1-\beta_t}}(y_t+\beta_is_\theta(y_t,t,x),\beta_tI),
\end{equation}
where the objective is proved equivalent to the following by score matching~\cite{song2019generative}:
\begin{equation}
    \label{fisher}
    \mathbb{E}_{p(x,y)}[||s_\theta(y_t,t,x)-\nabla_{y_t}\log p(y_t|y_0)||_2^2].
\end{equation}

After training the score-based model DiffIML $s_\theta$ by minimizing the Fisher divergence between the model and the actual data distribution as shown in Eq.(\ref{fisher}), we can generate masks by starting from $y_T \sim \mathcal{N}(0,I)$ and follow the estimated reverse Markov chain as below:
\begin{equation}
\label{chain}
\begin{split}
    y_{t-1}=\frac{1}{\sqrt{1-\beta_t}}(y_t+\beta_ts_\theta(y_t,t,x))+\sqrt{\beta_t}\epsilon, \\ \quad t=T,T-1,\cdots,1.
\end{split}
\end{equation}

The formulation above describes the generic diffusion process on the target data $y$. In our implementation, to reduce computational costs, DiffIML performs this process in the latent space (replacing $y$ with latent $z$), as detailed in the following section.

It can be observed that two bottlenecks exist within the aforementioned mechanism. First, as shown in Eq.(\ref{fisher}), the model estimates the gradient of the log probability (score) of the mask $y$ at each noise scale. Yet performing the iterative reverse process directly on the high-resolution pixel-wise mask $y$ incurs prohibitive computational costs and high latency. Secondly, error accumulates during the reverse process, causing inaccurate mask prediction. This is because the \textbf{\textit{prediction error}} of $s_\theta$ is unavoidable, causing \textbf{\textit{iterative error}} between the predicted $\hat{y}_{t-1}$ and the actual $y_{t-1}$ using Eq.(\ref{chain}). The iterative error leads to a greater prediction error in the subsequent step, as the model's input during training is $y_{t-1}$ rather than $\hat{y}_{t-1}$.

\subsection{Model Architecture for Efficiency}
As illustrated in Figure \ref{fig:network}, DiffIML employs a dual-efficiency architecture comprising a Lightweight Mask-Specific VAE (LightVAE) as the latent encoder and a decoupled denoising pipeline to ensure real-time inference.

\textbf{LightVAE.} We first validate that the pre-trained VAE~\cite{vae} from Stable Diffusion~\cite{stablediff}, despite being designed for RGB images, enables effective latent modeling for both binary masks and edge maps in IML. Specifically, although the masks and edge maps are single-channel, we observe that the reconstruction error of them using the VAE is negligible by duplicating into three identical channels, \textit{i.e.}, $y \approx \mathcal{D}(\mathcal{E}(y))$, confirming that the pre-trained latent space preserves structural details effectively. Therefore, we perform the proposed DiffIML on this robust latent representation $z^{(y)}=\mathcal{E}(y)$. 

However, VAE in Stable Diffusion is computationally expensive. To retain only its mask reconstruction capability, we introduce a Lightweight Mask-Specific VAE distilled from the VAE in Stable Diffusion. LightVAE removes heavy attention mechanisms and adopts a streamlined fully convolutional structure. It compresses the mask into a compact latent space with $1/8$ resolution via simple downsampling blocks, minimizing inference latency. Visualized as the Latent Encoder in Figure \ref{fig:network}. Specifically, we utilize a frozen pre-trained VAE in Stable Diffusion as the teacher $(\mathcal{E}_{T}, \mathcal{D}_{T})$ and our LightVAE as the student $(\mathcal{E}_{S}, \mathcal{D}_{S})$. The student is trained to minimize a dual objective:
\begin{equation}
   \mathcal{L}_{distill} = \| \mathcal{D}_S(z_S) - y \|_1 + \lambda_{lat} \| z_S - z_T \|_2^2,
\end{equation}
\noindent where $y$ denotes the ground truth binary mask, and $z_S = \mathcal{E}_S(y)$ is the latent code encoded by the student. $z_T = \mathcal{E}_T([y, y, y])$ represents the target latent extracted by the frozen teacher, where $[y, y, y]$ denotes duplicating the single-channel mask along the channel dimension to satisfy the teacher's RGB input requirement. $\lambda_{lat}$ is a balancing hyperparameter. 
By replacing the heavy teacher with the student during inference, LightVAE reduces the encoding-decoding latency from 146ms to 1.3ms, effectively mitigating the computational bottleneck.

\textbf{Decoupled Denoising Pipeline.} We implement a decoupled denoising strategy to resolve the efficiency limitations of standard iterative diffusion. This pipeline separates the static feature extraction from the dynamic sampling process. For static feature extraction, the UNet uses artifacts as the condition to accurately locate the tampered regions. However, artifacts are difficult to describe and cannot be labeled by humans~\cite{mvsspp_2022, trufor2023, NCL_IML_2023}. Therefore, we use a trainable condition encoder to learn to extract artifacts. This condition encoder is flexible and can be replaced with any backbone, such as ResNet~\cite{resnet} or Segformer~\cite{xie2021segformer}. Then, the computationally expensive iterative denoising is delegated to a simplified UNet with restricted channel depth.

\begin{algorithm}[t]
    \caption{DiffIML Training}
    \Description{Training procedure of DiffIML. The algorithm extracts image conditions and latent representations of the mask and edge map, adds diffusion noise with an error prior, predicts latent outputs with a U-Net, splits mask and edge latents, computes segmentation and edge losses, and returns the weighted combined loss.}
    \label{al_train}
    \begin{algorithmic}[1]
        \STATE \textit{\textbf{def} train(image, mask, edge):}
        \STATE \quad \# Extract conditions and latents
        \STATE \quad $c = condition\_encoder(image) $
        \STATE \quad $z_y = student\_encoder(mask)$
        \STATE \quad $z_e = student\_encoder(edge)$
        \STATE \quad $z_0 = Concat([z_y, z_e])$
        \STATE \quad \# Noisy and prior $\hat{z}_t$
        \STATE \quad $t \sim \mathcal{U}(\{1, \dots, T\}), \epsilon \sim \mathcal{N}(0, I), \xi \sim \mathcal{N}(0, I)$
        \STATE \quad $\hat{z}_t = \sqrt{\bar{\alpha}_t}(z_0 + \lambda \xi) + \sqrt{1 - \bar{\alpha}_t}\epsilon$
        \STATE \quad \# Predict latents
        \STATE \quad $z_{pred} = unet(\hat{z}_t, t, c)$
        \STATE \quad \# Combined loss
        \STATE \quad $z^{(y)}_{pred}, z^{(e)}_{pred} = split(z_{pred})$
        \STATE \quad $loss_{seg} = mse\_loss(z^{(y)}_{pred}, z_y)$
        \STATE \quad $loss_{edg} = mse\_loss(z^{(e)}_{pred}, z_e)$
        \STATE \quad $loss = \alpha * loss_{seg} + (1 - \alpha) * loss_{edg}$
        \STATE \quad \textit{\textbf{return} loss}
    \end{algorithmic}
\end{algorithm}

\begin{algorithm}[t]
    \caption{DiffIML Inference}
    \Description{Inference procedure of DiffIML. The algorithm extracts the image condition once, performs multiple rounds of latent diffusion sampling and iterative denoising, averages the resulting latent representations, decodes the mask latent, and returns the final predicted mask.}
    \label{al_sample}
    \begin{algorithmic}[1]
        \STATE \textit{\textbf{def} infer(image, infer\_steps, avg\_times)}:
        \STATE \quad \# Extract condition once
        \STATE \quad $c = condition\_encoder(image)$
        \STATE \quad \# Ensemble sampling
        \STATE \quad \textit{$z_{total} = 0$}
        \STATE \quad \textit{for i in range(avg\_times):}
        \STATE \quad \quad $z_t \sim \mathcal{N}(0, I)$
        \STATE \quad \quad \# Iteratively denoising
        \STATE \quad \quad \textit{for t in reverse range(infer\_steps):}
        \STATE \quad \quad \quad $z_{pred} = unet(z_t, t, c)$
        \STATE \quad \quad \quad $z_{t-1} \sim q(z_{t-1}|z_t, z_{pred})$
        \STATE \quad \quad \quad $z_t = z_{t-1}$
        \STATE \quad \quad $z_{total} = z_{total} + z_t$
        \STATE \quad \# Decode final mask
        \STATE \quad $z_{avg} = z_{total} / avg\_times$
        \STATE \quad $z^{(y)}_{avg}, z^{(e)}_{avg} = split(z_{avg})$
        \STATE \quad $y = student\_decoder(z^{(y)}_{avg})$
        \STATE \quad \textit{\textbf{return} $y$}
        \end{algorithmic}
\end{algorithm}

\subsection{Edge Supervision and Error Prior for Stability}

The second challenge in score-based IML is the stability bottleneck caused by error accumulation during the iterative reverse process. As previously mentioned, error accumulation at step $t$ is caused by two factors: (i) prediction error; (ii) iterative error between $\hat{z}_{t-1}$ and $z_{t-1}$. Therefore, we propose edge supervision and error prior to alleviating prediction error and iterative error, respectively.

\textbf{Edge Supervision.} Manipulating images often struggles to maintain consistency between tampered and authentic regions, thus leaving artifacts along the edges. Therefore, we introduce edge supervision to ensure that the model pays additional attention to edge information to better capture edge artifacts, thereby reducing prediction error. We generate the ground truth edge map $e$ from the binary mask $y$ using the Canny algorithm followed by a dilation operation:
\begin{equation}
    e=dilate(Canny(y))
\end{equation}

As shown in Algorithm~\ref{al_train}, we introduce an auxiliary edge loss to better capture edge artifacts. The edge map $e$ is encoded by the student encoder $\mathcal{E}_S$ into the latent edge representation $z^{(e)} = \mathcal{E}_S(e)$. The UNet is tasked to reconstruct both the mask latent $z^{(y)}$ and the edge latent $z^{(e)}$ simultaneously. The training objective is composed of the mask segmentation loss and the auxiliary edge loss:
\begin{equation}
    \mathcal{L}_{seg} = \| z^{(y)} - \hat{z}^{(y)}_{pred} \|_2^2 \\
\end{equation}
\begin{equation}
    \mathcal{L}_{edg} = \| z^{(e)} - \hat{z}^{(e)}_{pred} \|_2^2 \\
\end{equation}
where $\hat{z}^{(y)}_{pred}$ and $\hat{z}^{(e)}_{pred}$ are the predictions from the UNet. $z^{(y)}$ and $z^{(e)}$ are the latent representations. $\mathcal{L}_{seg}$ directly controls the estimate of the score function, while $\mathcal{L}_{edg}$ assists $\mathcal{L}_{seg}$ in learning better boundary details. We use a combined loss of the two losses:
\begin{equation}
    \mathcal{L}_{total} = \alpha \mathcal{L}_{seg} + (1-\alpha) \mathcal{L}_{edg}
\end{equation}
where $\alpha \in (0,1)$ is the segmentation weight. Both $\mathcal{L}_{seg}$ and $\mathcal{L}_{edg}$ are mean squared error (MSE) loss as they perform in latent space. We do not use the latent decoder to bring the prediction into pixel space and apply cross-entropy loss or dice loss~\cite{dice_loss} because this would significantly increase training time and reduce the precision of gradient propagation.

\textbf{Error Prior.} We treat iterative error as a prior during training, allowing the model to learn in advance, which means taking $\hat{z_t}$ as the input during training rather than $z_t$. However, coming from the sampling chain, $\hat{z_t}$ is unknown in the training phase. Inspired by previous works~\cite{bao2022analytic}, we model $\hat{z_t}$ as $p_\theta(\hat{z_t}|z_t)$ and approximate it using a Gaussian distribution:
\begin{equation}
    p_\theta(\hat{z_t}|z_t)=\mathcal{N}(\hat{z_t};z_t,\zeta_t^2I)
\end{equation}
\begin{equation}
    \hat{z_t}=z_t+\zeta_t*\xi \quad (\xi\sim\mathcal{N}(0,I))
    \label{ep1}
\end{equation}

From $q({y}_{t}|{y}_{0})=\mathcal{N}\left({y}_{t};\sqrt{\bar{\alpha}_{t}} {y}_{0},\left(1-\bar{\alpha}_{t}\right) {I}\right)$, we have:
\begin{equation}
    z_t = \sqrt{\bar{\alpha}_t} z_0 + \sqrt{(1-\bar{\alpha}_t)} \epsilon \quad (\epsilon\sim\mathcal{N}(0,I))
    \label{eb2}
\end{equation}

Combining Eq. \ref{ep1}, Eq. \ref{eb2}, we have:
\begin{equation}
    \hat{z_t} = \sqrt{\bar{\alpha}_t} z_0 + \sqrt{(1-\bar{\alpha}_t)} \epsilon+\zeta_t*\xi\quad (\epsilon,\xi\sim\mathcal{N}(0,I))
    \label{eb3}
\end{equation}

Selecting the best schedule $(\zeta_T,...,\zeta_0)$ demands a substantial amount of time to finetune. Intuitively, $\zeta_t$ increases as step $t$ decreases because iterative error accumulates over the steps. Hence, we have opted for a simple yet effective solution: setting $\zeta_t=\sqrt{\bar{\alpha}_t} \lambda $. The constant $\lambda$ is called prior rate. Eq. \ref{eb3} thus transforms into:

\begin{equation}
    \hat{z_t} = \sqrt{\bar{\alpha}_t} (z_0 + \lambda \xi) + \sqrt{(1-\bar{\alpha}_t)} \epsilon \quad (\epsilon,\xi\sim\mathcal{N}(0,I))
    \label{eb4}
\end{equation}

During training, we use Eq. \ref{eb4} instead of Eq. \ref{eb2} to generate the model's input, as shown in Algorithm~\ref{al_train}. During generation, the diffusion formula is not affected by the proposed error prior. Therefore, the introduction of the error prior is simple to implement and plug-in.

\subsection{Generation}

As shown in Algorithm~\ref{al_sample}, the condition encoder runs once to obtain the condition given a test image during generation. The model samples a random noise from a Gaussian distribution and then iteratively refines it. Ultimately, we compute the average of the sampled latents and decode it using the student decoder to obtain the final predicted mask.

\textbf{Denoising Rule.} We follow the DDIM~\cite{ddim} rule to perform respaced steps for faster denoising. The $\sigma$ of DDIM is set to 0 to make the sampling chain determined because one image only corresponds to a unique mask in IML. Unlike in generation tasks, DiffIML does not require many sampling steps to achieve good performance. This is because, compared to reconstructing an RGB image, generating a binary mask does not need many steps to accumulate information, which happens to be extremely beneficial for DiffIML's inference efficiency.

\textbf{Ensemble Generation.} Ideally, given an image $x$ as a condition, DiffIML’s generated distribution should consist of a unique mask $y$. In this case, even though the diffusion process starts from random Gaussian noise, DiffIML always finds a generation path that results in a final target distribution containing only $y$. However, in practice, the distribution generated by DiffIML may deviate from $y$, so we run the denoising process $N$ times given one image condition and generate an ensemble of masks. This method enhances both prediction accuracy and stability. Additionally, ensemble only increases the number of execution of the lightweight UNet, so we can control the size of $N$ to achieve a trade-off between accuracy, stability, and generation speed. 

\textbf{LightVAE Reconstruction Quality.} We evaluated the reconstruction upper bound. The reconstruction yields a pixel-level F1 score of 0.9970 (at a threshold of 0.5).
\section{Experiments}

\subsection{Datasets} 
Consistent with methods~\cite{salloum2018image,zhou2018learning,mvss,mvsspp_2022,kong2023pixel,ma2023iml,forensichub}, we primarily adopt the most widely used MVSS protocol~\cite{ma2025imdl} to assess the generalization of DiffIML. In this protocol, the model is trained only on CASIAv2~\cite{casiav1} and then tested directly on other datasets without fine-tuning. 
Furthermore, to demonstrate the model's superior generalization capability under a large-scale dataset, we additionally employ the CAT-Net protocol~\cite{CAT-Net2022}.

\begin{table}[t]
    \centering
    \newcommand{\TrMark}{\scalebox{1.0}{\textcolor{blue}{$\bigcirc$}}} 
    \newcommand{\TeMark}{\scalebox{1.3}{\textcolor{red}{$\bigtriangleup$}}}

    \caption{Summary of image manipulation datasets used in MVSS and CAT-Net protocols. Au and Tp represent authentic and tampered images. \TrMark~denotes Training set, \TeMark~indicates Test set.}
    \Description{A summary table of the datasets used in the MVSS and CAT-Net protocols for image manipulation localization. For each dataset, the table lists the year, the numbers of authentic and tampered images, whether the manipulation is DGM-based, the manipulation pipeline type, and whether the dataset is used for training or testing under each protocol.}
    \label{tab:dataset_merged}
    
    \resizebox{\linewidth}{!}{
        \begin{tabular}{l c cc c l cc}
            \toprule
            \multirow{2}{*}{\textbf{Dataset}} & \multirow{2}{*}{\textbf{Year}} & \multirow{2}{*}{\textbf{Au}} & \multirow{2}{*}{\textbf{Tp}} & \multirow{2}{*}{\textbf{DGM-based}} & \multirow{2}{*}{\textbf{Pipeline}} & \multicolumn{2}{c}{\textbf{Protocol}} \\
            \cmidrule(lr){7-8}
             & & & & & & \textbf{MVSS} & \textbf{CAT} \\
            \midrule
            
            CASIAv2~\cite{casiav1} & 2013 & 7491 & 5063 & $\times$ & Manual & \TrMark & \TrMark \\
            
            FantasticReality~\cite{fantastic_reality} & 2019 & 16592 & 19423 & \checkmark & Random & -- & \TrMark \\
            tamperedCOCO~\cite{CAT-Net2022} & 2022 & 0 & 799441 & $\times$ & Random & -- & \TrMark \\
            compRAISE~\cite{CAT-Net2022} & 2022 & 24462 & 0 & $\times$ & Random & -- & \TrMark \\
            
            CASIAv1~\cite{casiav1} & 2013 & 800 & 920 & $\times$ & Manual & \TeMark & \TeMark \\
            COVERAGE~\cite{cover} & 2016 & 100 & 100 & $\times$ & Manual & \TeMark & \TeMark \\
            Columbia~\cite{columbia} & 2006 & 183 & 180 & $\times$ & Random & \TeMark & \TeMark \\
            NIST16~\cite{nist16} & 2019 & 0 & 564 & $\times$ & Manual & \TeMark & \TeMark \\
            
            IMD2020~\cite{IMD20_2020} & 2020 & 414 & 2010 & $\times$ & Manual & \TeMark & \TrMark \\
            
            CocoGlide~\cite{trufor2023} & 2023 & 0 & 512 & \checkmark & Random & \TeMark & \TeMark \\
            Autosplice~\cite{jia2023autosplice} & 2023 & 0 & 3621 & \checkmark & Manual & \TeMark & \TeMark \\
            
            \bottomrule
        \end{tabular}
    }
\end{table}
Additionally, considering recent region-editing manipulations based on deep generative models, we include two generative datasets, CocoGlide~\cite{trufor2023} and Autosplice~\cite{jia2023autosplice}, for testing. The detailed statistics of all training and testing datasets are summarized in Table \ref{tab:dataset_merged}.
\begin{table*}[h] 
    \centering
    \caption{\textbf{Pixel-level F1 Performance of Image Manipulation Localization under the MVSS protocol.} The threshold is fixed at 0.5. Most of the F1 scores are from previous methods~\cite{mvss,mvsspp_2022,kong2023pixel,ma2023iml}, while the missing ones on certain datasets are from our re-implemented training. The best and second-best performances are highlighted in \textbf{bold} and \underline{underlined}.}
    \Description{A comparison table of pixel-level F1 scores under the MVSS protocol across seven test datasets and the average score. The methods are evaluated on both non-generative and generative datasets. DiffIML achieves the best average F1 score and obtains the best performance on most datasets, while OpenSDI achieves the best result on IMD2020.}
    \label{tab:com_v2}
    \resizebox{0.7\textwidth}{!}{
    \begin{tabular}{lcccccccc}
    \toprule
    \multirow{2}{*}{\textbf{Method}} & \multicolumn{5}{c}{\textbf{Non-generative}} & \multicolumn{2}{c}{\textbf{Generative}} & \multirow{2}{*}{\textbf{Average}}\\
    \cmidrule(r){2 - 6} \cmidrule(r){7 - 8}
    & COVERAGE & Columbia & NIST16 & CASIAv1 & IMD2020 & CocoGlide & Autosplice \\
    \midrule
    MVSS-Net\scriptsize{(\textit{ICCV21})} & \underline{0.453} & 0.638 & 0.292 & 0.452 & 0.260 & 0.291 & 0.298 & 0.383 \\
    CAT-Net\scriptsize{(\textit{IJCV22})}& 0.210 & 0.206 & 0.102 & 0.237 & 0.257 & 0.293 & 0.357 & 0.237 \\
    Trufor\scriptsize{(\textit{CVPR23})} & 0.248 & 0.851 & 0.218 & 0.602 & 0.249 & 0.205 & 0.390 & 0.395 \\
    NCL-IML\scriptsize{(\textit{ICCV23})} & 0.225 & 0.446 & 0.260 & 0.502 & 0.237 & 0.287 & 0.187 & 0.306 \\
    IML-ViT\scriptsize{(\textit{Arxiv24})} & 0.410 & 0.780 & 0.331 & 0.721 & 0.327 & 0.237 & 0.349 & 0.451 \\
    Mesoscopic\scriptsize{(\textit{AAAI25})} & 0.326 & 0.726 & 0.343 & \underline{0.740} & 0.269 & 0.162 & 0.249 & 0.402 \\
    SparseViT\scriptsize{(\textit{AAAI25})} & 0.305 & \underline{0.857} & 0.352 & 0.656 & 0.253 & 0.301 & \underline{0.413} & 0.448 \\
    OpenSDI\scriptsize{(\textit{CVPR25})} & 0.433 & 0.653 & \underline{0.402} & 0.695 & \textbf{0.513} & \underline{0.340} & 0.410 & \underline{0.492} \\
    \midrule
    DiffIML\scriptsize{(\textit{ours})} & \textbf{0.455} & \textbf{0.859} & \textbf{0.425} & \textbf{0.746} & \underline{0.417} & \textbf{0.380} & \textbf{0.452} & \textbf{0.533} \\
    \bottomrule
    \end{tabular}
    }
\end{table*}

\begin{table*}[h]
    \centering
    \caption{Pixel-level F1 Performance of Image Manipulation Localization under the CAT-Net protocol.}
    \Description{A comparison table of pixel-level F1 scores under the CAT-Net protocol across six test datasets and the average score. DiffIML achieves the best average performance and obtains the best results on most datasets, especially on the generative benchmarks.}
    \label{tab:com_catnet}
    \resizebox{0.7\textwidth}{!}{
    \begin{tabular}{lccccccc}
    \toprule
    \multirow{2}{*}{\textbf{Method}} & \multicolumn{4}{c}{\textbf{Non-generative}} & \multicolumn{2}{c}{\textbf{Generative}} & \multirow{2}{*}{\textbf{Average}}\\
    \cmidrule(r){2 - 5} \cmidrule(r){6 - 7}
    & COVERAGE & Columbia & NIST16 & CASIAv1 & CocoGlide & Autosplice \\
    \midrule
    MVSS-Net\scriptsize{(\textit{ICCV21})} & 0.482 & 0.740 & 0.336 & 0.583 & 0.443 & 0.385 & 0.495 \\
    CAT-Net\scriptsize{(\textit{IJCV22})}  & 0.427 & 0.915 & 0.252 & 0.808 & 0.410 & 0.387 & 0.533 \\
    Trufor\scriptsize{(\textit{CVPR23})}   & 0.457 & 0.885 & 0.348 & 0.818 & 0.283 & 0.393 & 0.531 \\
    NCL-IML\scriptsize{(\textit{ICCV23})} & 0.320 & 0.657 & 0.315 & 0.570 & 0.326 & 0.302 & 0.415 \\
    IML-ViT\scriptsize{(\textit{Arxiv24})} & 0.581 & 0.915 & 0.324 & 0.752 & 0.369 & 0.407 & 0.558 \\
    Mesoscopic\scriptsize{(\textit{AAAI25})}& \underline{0.586} & 0.890 & \textbf{0.392} & \textbf{0.840} & 0.450 & 0.402 & 0.593 \\
    SparseViT\scriptsize{(\textit{AAAI25})} & 0.521 & \underline{0.937} & \underline{0.382} & 0.826 & 0.461 & 0.502 & \underline{0.605} \\
    OpenSDI\scriptsize{(\textit{CVPR25})}   & 0.437 & 0.871 & 0.359 & 0.783 & \underline{0.469} & \underline{0.512} & 0.572 \\
    \midrule
    DiffIML\scriptsize{(\textit{ours})} & \textbf{0.601} & \textbf{0.939} & 0.341 & \underline{0.828} & \textbf{0.592} & \textbf{0.612} & \textbf{0.652} \\
    \bottomrule
    \end{tabular}
    }
\end{table*}

\subsection{Evaluation Metrics.}

We evaluate SoTA models’ pixel-level detection performances using the F1 score, which is the most commonly used metric in IML. The F1 score is a harmonic mean between precision and recall. ~\citet{ma2025imdl} point out that for the IML task, AUC often reflects an overoptimistic metric while IoU fails to handle the extreme imbalance between negative and positive samples. Therefore, F1 score best reflects the model's detection performance. Unlike the best threshold of F1 score used in MVSS-Net~\cite{mvss} and Trufor~\cite{trufor2023}, we set the threshold to a fixed 0.5 for two reasons. Firstly, in real-world scenarios, it is impossible to know the optimal threshold for test images in advance. Secondly, using a fixed threshold helps to ensure fair comparisons.

\subsection{Implementation Details.} 

We implement our DiffIML using PyTorch~\cite{pytorch}. Training was performed on eight NVIDIA RTX 3090 GPUs for 200 epochs, with a total batch size of 128. The input size is $512\times512$. The condition encoder is a Segformer-b3~\cite{xie2021segformer} initialized with ImageNet-1k pre-trained weight. We adopt the standard latent scale factor of 0.18215 to align the student's latent distribution with the teacher's, ensuring robust feature representation. We also adopt a simplified U-Net with channel depths of 128 for the diffusion process. And for the UNet, we set the number of the diffusion step to 1000 with a linear schedule. AdamW~\cite{adamw} optimizer is employed with a cosine learning rate decay~\cite{cosine_decay} from $10^{-4}$ to $0$ and weight decay of $0.05$. The segmentation weight $\alpha$ is set to $0.2$ to guide the model more focus on the edge region, and the prior rate $\lambda$ is set to $0.1$. During inference, we use 8 denoising steps and set the ensemble size N to 5 by default. For the main performance results in Table \ref{tab:com_v2} and Table \ref{tab:com_catnet}, we additionally apply a simple test-time augmentation (TTA) by horizontally flipping the input image, predicting both the original and flipped versions, and fusing the two predictions after flipping back. Following MVSS-Net~\cite{mvss}, we apply data augmentation for training, including flipping, blurring, compression, and naive manipulations either by cropping and pasting a squared area or using built-in OpenCV inpainting functions.

\subsection{Generalization Comparison with SoTA}
\label{sec:sota}
\begin{table}[h]
    \centering
    \caption{Comparison of Params, FLOPs and FPS.}
    \Description{A comparison table of model efficiency in terms of input size, parameter count, FLOPs, and inference speed. DiffIML with LightVAE greatly reduces parameters and FLOPs compared with the VAE version, while achieving competitive real-time FPS at 512 by 512 resolution.}
    \label{tab:efficiency}
    \resizebox{0.48\textwidth}{!}{
    \begin{tabular}{lcccc}
    \toprule
    \textbf{Method} & \textbf{Size} & \textbf{Params (M)} & \textbf{FLOPs (G)} & \textbf{FPS} \\
    \midrule
    MVSS-Net \scriptsize{(ICCV'21)} & $512\times512$ & 146.80 & 166.70 & 41.27 \\
    Trufor \scriptsize{(CVPR'23)}   & $512\times512$ & 68.70 & 236.50 & 36.22 \\
    CAT-Net \scriptsize{(IJCV'22)}  & $512\times512$ & 114.30 & 134.00 & 39.17 \\
    PSCC-Net \scriptsize{(TCSVT'21)} & $256\times256$ & 3.70 & 45.30 & 120.75 \\
    ManTraNet \scriptsize{(CVPR'19)} & $256\times256$ & 3.90 & 274.00 & 61.11 \\
    ObjectFormer \scriptsize{(CVPR'22)} & $224\times224$ & 131.84 & 246.70 & 102.58 \\
    IML-ViT \scriptsize{(ICCV'23)}  & $1024\times1024$ & 91.74 & 136.22 & 19.53 \\
    DiffIML \scriptsize{(VAE, N=1)} & $512\times512$ & 132.10 & 213.20 & 21.92 \\
    DiffIML \scriptsize{(LightVAE, N=1)} & $512\times512$ & 48.51 & 53.01 & 69.48 \\
    DiffIML \scriptsize{(LightVAE, N=5)} & $512\times512$ & 48.51 & 102.36 & 37.28 \\
    \bottomrule
    \end{tabular}
    }
\end{table}
The results are shown in Table \ref{tab:com_v2}. Most of the F1 scores are from previous methods~\cite{mvss,mvsspp_2022,kong2023pixel,ma2023iml}, while the missing ones on certain datasets are from our re-implemented training, such as the generative datasets. DiffIML achieves the best results on six out of the seven test datasets and performs well on the remaining dataset. 

It can be observed that previous methods, such as Mesoscopic, perform well on CASIAv1, which shares the same source as the training set CASIAv2, but show poor generalization on other datasets. Although the performance on IMD2020 is slightly lower than OpenSDI, DiffIML demonstrates a more balanced and robust generalization ability overall. Compared to the sub-optimal OpenSDI, DiffIML improves the average F1 score from 0.492 to 0.533, corresponding to an 8.3\% relative improvement. The superiority of DiffIML stems from the better generalization of the score-based model in estimating the gradient of the data distribution, rather than memorizing specific tampering artifacts.

To provide a more holistic assessment of model robustness, particularly against diverse manipulation techniques, we also adopt the CAT-Net protocol~\cite{CAT-Net2022}.
It shows that our dominance on generative datasets highlights the diffusion paradigm's efficacy in detecting subtle high-frequency artifacts. Moreover, superior performance on hard benchmarks such as COVERAGE, together with strong results on generative datasets, confirms the model’s robustness in diverse and multi-source environments.

\subsection{Efficiency Analysis}

\begin{table}[t]
    \centering
    \small
    \caption{Effectiveness of the generative diffusion paradigm under the MVSS protocol.}
    \Description{A comparison table of pixel-level F1 scores under the MVSS protocol between a discriminative SegFormer-B3 baseline and the proposed DiffIML. DiffIML outperforms SegFormer-B3 on all non-generative and generative test datasets, achieving a higher average F1 score.}
    \label{tab:baseline}
    \setlength{\tabcolsep}{1pt}
    \begin{tabular}{lcccccccc}
    \toprule
    \multirow{2}{*}{\textbf{Method}} & \multicolumn{5}{c}{\textbf{Non-Generative}} & \multicolumn{2}{c}{\textbf{Generative}} & \multirow{2}{*}{\textbf{Average}} \\
    \cmidrule(lr){2-6} \cmidrule(lr){7-8}
    & \makecell{\textbf{Colu-}\\\textbf{mbia}} & \makecell{\textbf{COVE-}\\\textbf{RAGE}} & \makecell{\textbf{NIST}\\\textbf{16}} & \makecell{\textbf{CASIA}\\\textbf{v1}} & \makecell{\textbf{IMD}\\\textbf{2020}} & \makecell{\textbf{Auto-}\\\textbf{splice}} & \makecell{\textbf{Coco-}\\\textbf{Glide}} & \\
    \midrule
    SegFormer-B3 & 0.806 & 0.394 & 0.375 & 0.557 & 0.322 & 0.443 & 0.301 & 0.457 \\
    DiffIML\scriptsize{(\textit{Ours})} & \textbf{0.859} & \textbf{0.455} & \textbf{0.425} & \textbf{0.746} & \textbf{0.417} & \textbf{0.452} & \textbf{0.380} & \textbf{0.533} \\
    \bottomrule
    \end{tabular}
\end{table}
We evaluate the model efficiency in terms of Parameters, FLOPs, and FPS on a cluster of eight NVIDIA RTX 3090 GPUs. As shown in Table~\ref{tab:efficiency}, tested without Test-Time Augmentation (TTA), DiffIML achieves an inference speed of 69.48 FPS at $N=1$ and 37.28 FPS at $N=5$. In addition to speed, among the compared methods with $512 \times 512$ input resolution, DiffIML achieves the lowest FLOPs.
\begin{figure*}[t]
  \centering
  \includegraphics[width=0.88\linewidth]{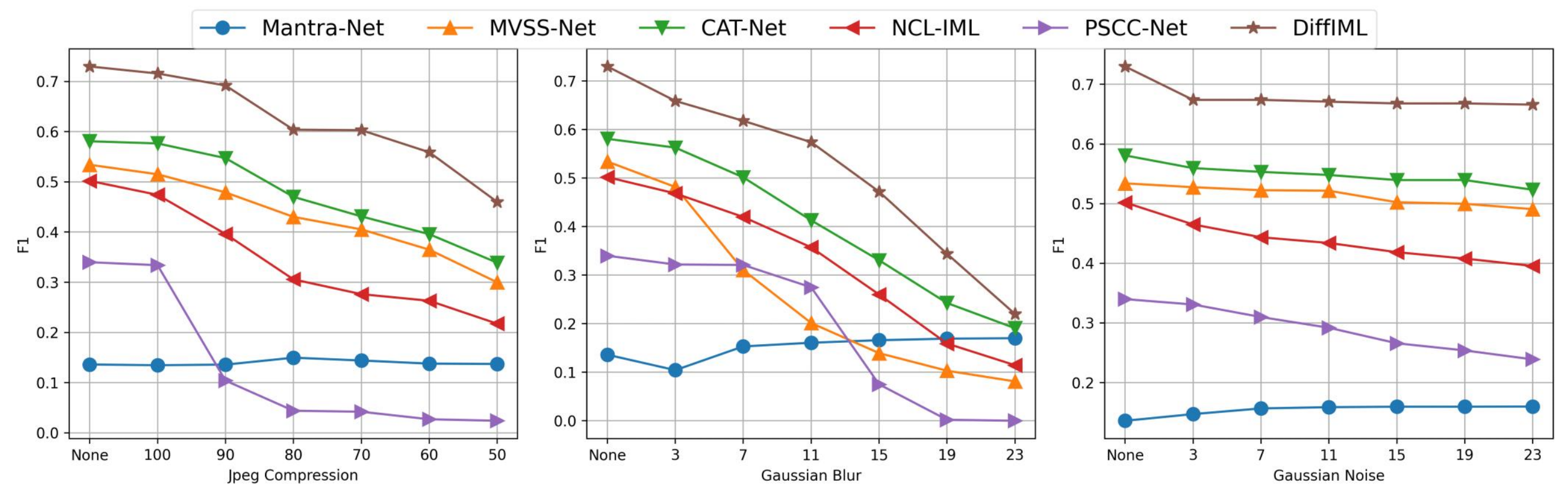}
  \captionsetup{skip=6pt}
  \caption{Results for robustness evaluation against Jpeg Compression, Gaussian Blur, and Gaussian Noise.}
  \Description{A figure showing robustness evaluation results of different methods under three common post-processing perturbations: JPEG compression, Gaussian blur, and Gaussian noise. The plots compare how model performance changes as the degradation level increases, showing that DiffIML maintains relatively stronger robustness under these distortions.}
  \label{fig:robustness}
\end{figure*}
\begin{table*}[thb] \centering
    \caption{\textbf{Ablation experiments} with training on CASIAv2 and testing on CASIAv1. The default settings for the 5 ablation experiments are as follows: using pre-trained weights, employing Segformer-b3 as the backbone for the condition encoder, setting the prior rate to 0.1, using edge supervision, and utilizing 8 sampling steps.}
    \Description{An ablation table with five parts evaluated by F1 score on CASIAv1 after training on CASIAv2. Part (a) compares initialization methods and shows ImageNet-1k pretraining outperforms Xavier initialization. Part (b) compares condition encoder backbones and shows Segformer-b3 performs best. Part (c) compares prior rates and shows 0.1 is the best setting. Part (d) compares using edge supervision or not and shows edge supervision improves performance. Part (e) compares different sampling steps, where 10 steps gives the highest F1, while 8 steps provides the best trade-off between performance and efficiency.}
    \label{tab:ablation}
    
    \resizebox{0.92\textwidth}{!}{
            { \footnotesize
            \begin{minipage}{0.14\textwidth}
                \begin{tabular}{c|c}
                    \toprule
                    Init & F1 \\
                    \midrule
                    Xavier & 0.412 \\
                    \textbf{IN-1K} & \textbf{0.746}\\
                    \bottomrule
                \end{tabular}
            \end{minipage}
            }
            \hspace{0.02\textwidth}

            { \footnotesize
            \begin{minipage}{0.16\textwidth}
                \begin{tabular}{c|c}
                    \toprule
                    Backbone & F1 \\
                    \midrule
                    Res-50 & 0.542 \\
                    Res-101 & 0.653 \\
                    Seg-b2 & 0.624 \\
                    \textbf{Seg-b3} & \textbf{0.746} \\
                    Seg-b4 & 0.711 \\
                    \bottomrule
                \end{tabular}
            \end{minipage}
            }
            
            \hspace{0.02\textwidth}

            { \footnotesize
            \begin{minipage}{0.14\textwidth}

                \begin{tabular}{c|c}
                    \toprule
                    Prior & F1 \\
                    \midrule
                    0.0 & 0.713 \\
                    \textbf{0.1} & \textbf{0.746} \\
                    0.3 & 0.706 \\
                    0.5 & 0.659 \\
                    \bottomrule
                \end{tabular}
            \end{minipage}
            }
            
            \hspace{0.02\textwidth}

            { \footnotesize
            \begin{minipage}{0.14\textwidth}
                \begin{tabular}{c|c}
                    \toprule
                    Edge & F1 \\
                    \midrule
                    $\times$ & 0.649 \\
                    \textbf{\checkmark} & \textbf{0.746} \\
                    \bottomrule
                \end{tabular}
            \end{minipage}
            }
            
            \hspace{0.02\textwidth}

            { \footnotesize
            \begin{minipage}{0.14\textwidth}
                \begin{tabular}{c|c}
                    \toprule
                    Step & F1 \\
                    \midrule
                    1 & 0.681 \\
                    5 & 0.708 \\
                    8 & 0.746 \\
                    \textbf{10} & \textbf{0.749} \\
                    \bottomrule
                \end{tabular}
            \end{minipage}
            }

    }
    
    \resizebox{0.95\textwidth}{!}{
        \begin{minipage}{\textwidth}
        \centering
            \begin{minipage}[t]{0.174\textwidth}
                {\footnotesize (a) \textbf{Init}. The init of pretrained ImageNet-1k weights works better.}
            \end{minipage}
            \hspace{0.012\textwidth}
            \begin{minipage}[t]{0.204\textwidth}
                {\footnotesize (b) \textbf{Condition encoder}. The best backbone for condition encoder is Segformer-b3.}
            \end{minipage}
            \hspace{0.01\textwidth}
            \begin{minipage}[t]{0.144\textwidth}
                {\footnotesize (c) \textbf{Prior rate}. The best prior rate is 0.1.}
            \end{minipage}
            \hspace{0.01\textwidth}
            \begin{minipage}[t]{0.194\textwidth}
                {\footnotesize (d) \textbf{Edge supervision}. Edge supervision works better.}
            \end{minipage}
            \hspace{0.01\textwidth}
            \begin{minipage}[t]{0.184\textwidth}
                {\footnotesize (e) \textbf{Sampling steps}. For both performance and efficiency, the 8 step is the best.}
            \end{minipage}
        \end{minipage}
    }

\end{table*}

\subsection{Ablation Study}

We train our DiffIML on CASIAv2~\cite{casiav1} and evaluate the pixel F1 score on CASIAv1~\cite{casiav1}. Results are shown in Table \ref{tab:baseline} , \ref{tab:ablation} and \ref{tab:ensemble}.
\begin{table}[t]\centering
    \caption{\textbf{Ablation experiment of ensemble.} The F1 column represents the average value and standard deviation of five runs. The default setting is bolded.}
    \Description{An ablation table of ensemble size evaluated by F1 score, reported as mean and standard deviation over five runs. Increasing the ensemble size improves average performance and generally reduces variance, and the default setting of 5 provides a good trade-off between accuracy and stability.}
    \label{tab:ensemble}
    \resizebox{0.20\textwidth}{!}{
    \large
    \begin{tabular}{c|c}
        \toprule
       Ensemble & F1 \\
        \midrule
        1 & 0.7185 ± 0.0042 \\
        3 & 0.7392 ± 0.0028 \\
        \textbf{5} & \textbf{0.7461 ± 0.0015} \\
        10 & 0.7483 ± 0.0018 \\
        \bottomrule
    \end{tabular}
    }
\end{table}

\textbf{Generative Diffusion Paradigm.}
\label{sec:diff_paradigm}
Table \ref{tab:baseline} shows DiffIML significantly outperforms the deterministic SegFormer-B3 baseline, demonstrating the distinct advantage of the generative paradigm in capturing diverse manipulation traces.

\textbf{Initialization.} As shown in Table \ref{tab:ablation}(a), DiffIML without pre-trained initialization exhibits a significant decrease in performance.

\textbf{Condition Encoder.} The results show that accuracy first increases and then decreases with the increase in the number of parameters. Ultimately, Segformer-b3 and ResNet101, which have similar parameters in scale, achieve the best results.

\textbf{Prior Rate.} As shown in Table \ref{tab:ablation}(c), a prior rate of 0.1 performs better than not using a prior rate, while larger values lead to decreased performance. This is because, with larger rate values, the model's input distribution deviates more from the noising distribution of the diffusion than the error caused by error accumulation, resulting in reduced accuracy.

\textbf{Edge Supervision.} Table \ref{tab:ablation}(d) confirms that edge supervision improves performance by guiding the model to focus on tampering boundary artifacts.

\textbf{Sampling Steps.} Unlike in generation tasks, where models often require a great many steps for optimal results, in IML, our model achieves good accuracy even at 1 step. In comparison, the F1 score with 8 steps of sampling is 9.54\% higher than that with 1 step of sampling, demonstrating the effectiveness of the diffusion iterative mechanism in improving performance. Performance saturates beyond 8 steps. Thus, we select 8 for efficiency. 

\textbf{Ensemble.} Table \ref{tab:ensemble} shows the impact of ensemble generation on the accuracy and stability of DiffIML's prediction results. With the default setting, the average F1 score increases from 0.7185 to 0.7461, corresponding to a 3.8\% relative improvement, while the standard deviation decreases by 0.0027. To balance the trade-off with generation efficiency, we select 5 of ensemble as the default setting. 

\subsection{Robustness Test}

We conduct the robustness test on the CASIAv1 dataset to evaluate the accuracy changes of the models when faced with common post-processing techniques: JPEG Compression, Gaussian Blur, and Gaussian Noise. To reduce computational costs, all models in the robustness tests are trained only on the tampered images from CASIAv2. The pixel-level F1 results in Figure \ref{fig:robustness} indicate that our proposed DiffIML maintains high accuracy even under these commonly encountered real-world post-processing conditions. Notably, the diffusion mechanism, which involves continuously learning with Gaussian noise added to the model input, allows DiffIML to achieve relatively low accuracy degradation when handling image transformations with Gaussian Noise. This confirms that DiffIML succeeds by modeling the geometric properties of the distribution.
\section{Conclusion}

We propose DiffIML, the first score-based model designed for IML, which learns the more generalized geometric properties of the distribution by estimating the gradient of the log-likelihood. To ensure practicality, we introduce a Dual-Efficiency Architecture to resolve the latency bottleneck, alongside specific Error Mitigation Strategies to tackle error accumulation during iterative sampling. Extensive experiments demonstrate that DiffIML achieves state-of-the-art generalization while maintaining real-time inference speed.

\begin{acks}
This work was supported by the Fundamental Research Funds for the Central Universities (Sichuan University) under the Research Base Guidance Program (Grant No. SCU2026D013/26041K4068).
\end{acks}

\bibliographystyle{ACM-Reference-Format}
\bibliography{main} 

\end{document}